\documentclass[final]{cvpr}

\usepackage{times}
\usepackage{epsfig}
\usepackage{graphicx}
\usepackage{amsmath}
\usepackage{amssymb}

\usepackage[pagebackref=true,breaklinks=true,colorlinks,bookmarks=false]{hyperref}

\def\cvprPaperID{****} 
\def\confYear{CVPR 2021}

\begin{document}

\title{Face-GCN: A Graph Convolutional Network for 3D Dynamic Face Identification/Recognition}

\author{Konstantinos Papadopoulos,
Anis Kacem,
Abdelrahman Shabayek and
Djamila Aouada 
\\Interdisciplinary Centre for Security, Reliability and Trust (SnT)\\
University of Luxembourg, Luxembourg\\ Email: \small{\{\tt konstantinos.papadopoulos,anis.kacem,abdelrahman.shabayek,djamila.aouada\}@uni.lu}
}

\maketitle

\begin{abstract}
Face identification/recognition has significantly advanced over the past years. However, most of the proposed approaches rely on static RGB frames and on neutral facial expressions. This has two disadvantages. First, important facial shape cues are ignored. Second, facial deformations due to expressions can have an impact on the performance of such a method. In this paper, we propose a novel framework for dynamic 3D face identification/recognition based on facial keypoints. Each dynamic sequence of facial expressions is represented as a spatio-temporal graph, which is constructed using 3D facial landmarks. Each graph node contains local shape and texture features that are extracted from its neighborhood. For the classification/identification of faces, a Spatio-temporal Graph Convolutional Network (ST-GCN) is used. Finally, we evaluate our approach on a challenging dynamic 3D facial expression dataset.

\end{abstract}

\section{Introduction}

Face identification has been an active research topic in computer vision. It finds application in many fields, namely human computer interaction, virtual reality and information security. Nevertheless, the performance of such applications can be greatly affected by variation in pose, illumination, expression or self-occlusions.

Various face identification approaches have been proposed \cite{guillaumin2009you,taigman2015web,schwartz2010robust,zhao2018towards} in the literature. While most of them show impressive performance, there is one factor to be considered. The vast majority of these approaches relies on appearance features extracted from 2D RGB images. In most cases, appearance features can be adequately informative. However, such features suffer from sensitivity to head pose variations and illumination changes \cite{}. 


To address this, 3D-based approaches were introduced \cite{gilani2018learning,yu2017sparse,jribi2019se}. By using richer geometrical features provided by 3D scanners, 3D-based face identification methods have shown robustness in cases appearance features are inconsistent (e.g. variation in illumination). Despite their promising performance, most 3D-based approaches were developed under the assumption that faces are static. In real-life, however, our faces show continuous deformations in the form of facial expressions. While most of state-of-the-art methods focused on the neutralization of these facial expressions \cite{}, some recent studies \cite{kacem2020space,cheng20184dfab} showed that exploiting such deformations can bring additional signatures that may help in recognizing the identity of the persons. 


One common constraint when analysing 3D faces consists of putting all 3D face meshes in a point-to-point correspondence often considered as a prepossessing step. The goal of this step, usually referred as non-rigid registration of 3D faces, is to find correspondences between all the vertices of the face meshes. Several works in literature tried to solve this problem \cite{}. This problem is even more challenging in the presence of large face deformations due to facial expressions and/or identity variations \cite{}. Some attempts have been proposed in \cite{li2015towards,mian2008keypoint,wang2018learning} to avoid the non-rigid registration step but they still require a dense key-point matching step. 

In this paper, we propose to address the aforementioned issues by proposing a novel framework for 3D dynamic face identification combining non-registered 3D face shapes and the corresponding appearance features. First, we construct a spatio-temporal graph using 3D facial landmarks. The nodes of this graph contain both appearance and shape features extracted from the neighborhood of each landmark. The identification is performed using a modified Spatio-temporal Graph Convolutional Network (ST-GCN) \cite{yan2018spatial}. Such networks have proven to be robust in various fields (e.g. action recognition, DNA prediction, etc.) and have the advantage of jointly capturing the spatial structure of graphs and the temporal evolution of individual nodes.

\begin{figure*}[h]
    \centering
    \includegraphics[width=\linewidth]{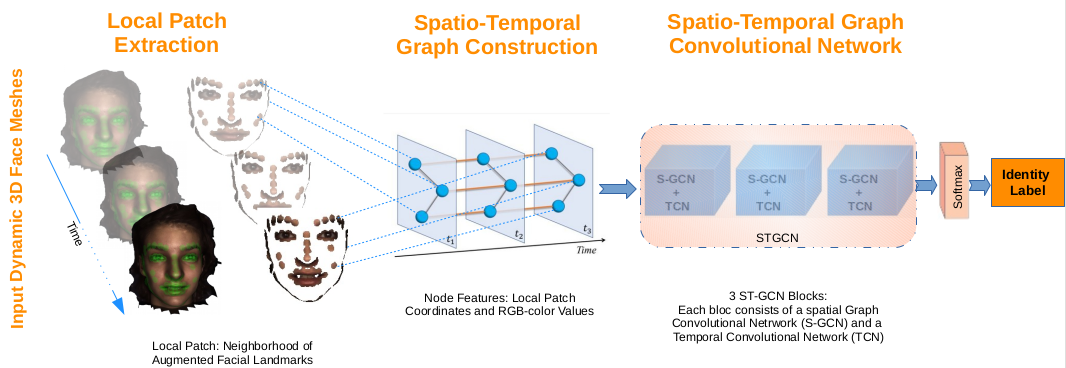}
    \caption{Face-GCN pipeline. Initially, facial landmarks are estimated from the 3D sequences of meshes and shape and texture features are extracted from their local patch. The next step is the construction of a spatio-temporal graph, in which the nodes contain the features from the local patches. The graph is then given as an input to the Spatio-Temporal Graph Convolutional Network. Best viewed in color.}
    \label{fig:overview}
\end{figure*}

The advantages of our approach are threefold. First, we propose a framework which considers the temporal factor in face identification, thus it is more suitable for real-life applications where people have a high variety of facial expressions. Second, the proposed framework does not require a 3D face non-rigid registration step which might introduce artefacts and distortions in the data. Finally, we exploit both shape and appearance features and use them within the same neural network for an effective early fusion of the two modalities. 


In summary, our contributions are the following:
\begin{itemize}
    \item A registration-free approach for dynamic 3D face identification, not requiring a dense point-to-point correspondence between 3D face meshes;
    \item The introduction of a GCN-based framework for dynamic 3D face identification;
    \item The introduction of local features for capturing both appearance and shape cues;
    \item An experimental validation and an extensive analysis of our approach on a challenging 3D dynamic facial expression dataset.
\end{itemize}

The structure of this paper is organized as follows: in Section~\ref{sec:related}, a literature review of the state-of-the-art approaches is presented. Section~\ref{sec:proposed} presents the proposed framework. The experiments and extensive analysis of the results are given in Section~\ref{sec:experiments}. Finally, Section~\ref{sec:conclusion} concludes this paper and discusses possible future directions.

\section{Related Work}
\label{sec:related}

In this section, we review important concepts on static and dynamic face identification and graph convolutional networks that are relevant to the context of this paper.

\paragraph{Static 3D Face Identification}

Over the last decades, 3D sensing has made notable advancements in computer vision. In particular, fields, such as face identification, in which facial shape information became available along with texture have gained significant popularity. For a more extensive survey of 3D face identification, one may consider \cite{soltanpour2017survey}.

There has been a wide variety of approaches on 3D static face identification \cite{kim2017deep,gilani2018learning,yu2017sparse,jribi2019se}. In \cite{kim2017deep}, an efficient method 3D face identification was presented. This approach relied on transfer learning from a CNN originally trained on 2D face images and fine-tuned on a relatively small number of 3D facial
scans. A different path was followed in \cite{gilani2018learning}, where a method for generating a large 3D face recognition dataset was introduced, along with a deep network trained exclusively on 3D faces. A storage-efficient 3D surface matching using 3D directional vertices was proposed in \cite{yu2017sparse}. This approach was applied on 3D face recognition and showed reduced computational time and storage requirements. Moreover, a 3D face description which is invariant under the Special Euclidean group SE(3) and independent to the original surface parameterization was proposed in \cite{jribi2019se}. 

\paragraph{Dynamic 3D Face Identification}

Static 3D face identification approaches have shown promising results and great potential for applications. However, in real life, faces are dynamic due to expressions, emotions, etc. Therefore, the development of models able to perform face identification in dynamic sequences is crucial. 

There is a wide variety of approaches on dynamic 3D face identification \cite{shao2017deep,kacem2020space, cheng20184dfab}. In \cite{shao2017deep}, a network learns robust dynamic texture information from fine-grained deep convolutional features to address 3D mask spoofing. A dynamic 3D verification approach was proposed in \cite{kacem2020space} in which a triplet loss network performed temporal modeling on encoded sequences of 3D faces. Furthermore, in \cite{cheng20184dfab}, a 3D dynamic face dataset was introduced for face recognition/verification. Experiments using temporal filtering and a simple LSTM network were also conducted as a reference.

\paragraph{Spatiotemporal Graph Convolutional Networks}

Graph convolutional networks (GCN) have been employed for a variety of tasks \cite{bruna2013spectral, henaff2015deep, duvenaud2015convolutional, li2015gated}. Recently, GCN made their way into skeleton-based action recognition \cite{yan2018spatial, li2019actional, shi2019two, shi2019skeleton, si2019attention, papadopoulos2019vertex} by exploiting the graph nature of human poses. One of the early approaches to apply GCN on skeleton sequences was proposed by Yan et al. \cite{yan2018spatial}. Skeleton sequences were converted into spatiotemporal graphs by preserving the inter-joint connections and linking the same joints from consecutive time steps temporally. The graph was then given as an input to a Spatio-temporal GCN, which consisted of multiple layers of graph convolutions. Two extensions were later proposed to capture more relevant dependencies between the joints \cite{li2019actional,shi2019two}. In \cite{li2019actional}, an autoencoder was used to capture action-specific and higher-order dependencies between joints, whereas in \cite{shi2019two} the joint connections were learned in an adaptive and end-to-end manner. In addition, directed spatiotemporal graphs were considered in \cite{shi2019skeleton} for better description of the relationship between joints and bones. Moreover, a more effective temporal modelling was proposed by \cite{papadopoulos2019vertex} and \cite{si2019attention}, where a hierarchical temporal modelling module and a Long Short-Term Memory network (LSTM) were respectively utilized.

\section{Proposed Approach}
\label{sec:proposed}

In this paper we propose a new framework for 3D dynamic face identification. The first step towards our approach is to construct patches defined by the neighborhoods of a set of facial landmarks. A spatio-temporal graph convolutional network is then used on different features extracted from the aforementioned patches along temporal dimension with the aim of extracting deformation signatures that are relevant for face identification.  


\subsection{3D Landmark Estimation}

The estimation of facial landmarks is very important in face analysis since they inform on the location of regions of interest, \emph{e.g.}, eye brows, mouth, nose, etc. While the estimation of these landmarks have been very accurate using 2D face images in literature~\cite{baltrusaitis2018openface,bulat2017far,zhu2016face}, less efforts have been made to estimate them in 3D face meshes. Accordingly, in this work we exploit the 2D texture face images and their mapping to 3D face meshes to estimate 3D facial landmarks. In particular, we use a robust state-of-the-art 2D facial landmark detector \cite{baltrusaitis2018openface} to estimate 2D landmarks, then we leverage these estimations to 3D using the texture mapping on 3D face meshes. Such mapping is generally provided in raw 3D scans acquired using recent 3D scanners. 

\begin{figure}[h]
    \centering
    \includegraphics[width=\linewidth]{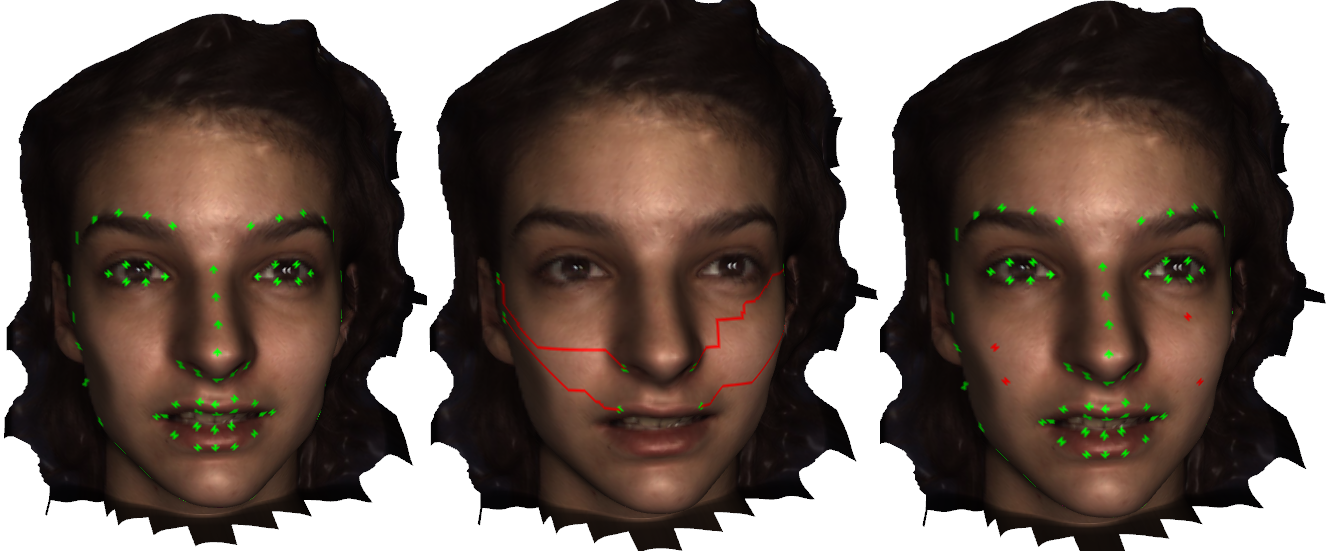}
    \caption{An example of estimated landmarks and their augmentation. From left to right: (1) Estimated landmarks by leveraging 2D estimations obtained using~\cite{zadeh2017convolutional}; (2) Geodesic paths between pairs of landmarks ; (3) Estimated landmarks (in red) and their augmentation (in green).Best viewed in color.}
    \label{fig:landmarks}
\end{figure}

In general, the number of estimated landmarks using state-of-the-art methods such as \cite{baltrusaitis2018openface,zhu2016face} can vary from $45$ to $90$ points detected around the eyes, eyebrows, nose, mouth, and chin. These landmarks describe well the face deformations and have been extensively used for face expression analysis \cite{jung2015joint,kacem2017novel}. However, they might be not sufficient in a face recognition scenario since they discard some important regions of the face such as the cheeks. To overcome these limitations, we augment the number of landmarks on the face by interpolating new points between the estimated landmarks. This is achieved by considering geodesic paths between pairs of landmarks using \cite{surazhsky2005fast} and taking the midpoints of these geodesic paths as new landmarks. Fig.~\ref{fig:landmarks} shows an example of estimated landmarks and their augmentation.

\subsection{Feature Extraction}
\label{sec:feature_extraction}

Face identification can be insufficient when only 3D facial landmarks are considered. Landmarks fail to describe discriminative facial patterns since crucial geometrical and texture information is omitted. Therefore, it is necessary to design features to describe these local facial patterns around landmarks and their evolution over time.

Given a 3D textured face mesh $\mathcal{M}$ and its full set of facial landmarks $i \in \mathbb{R}^3$, we propose to find a neighborhood $\mathcal{N}_i \subseteq \mathcal{M}$. For this purpose, we employ KD-tree search to find the closest points to each landmark. The final feature array is the concatenation of both local shape and texture patterns. 

\begin{figure}[h]
    \centering
    \includegraphics[width=\linewidth]{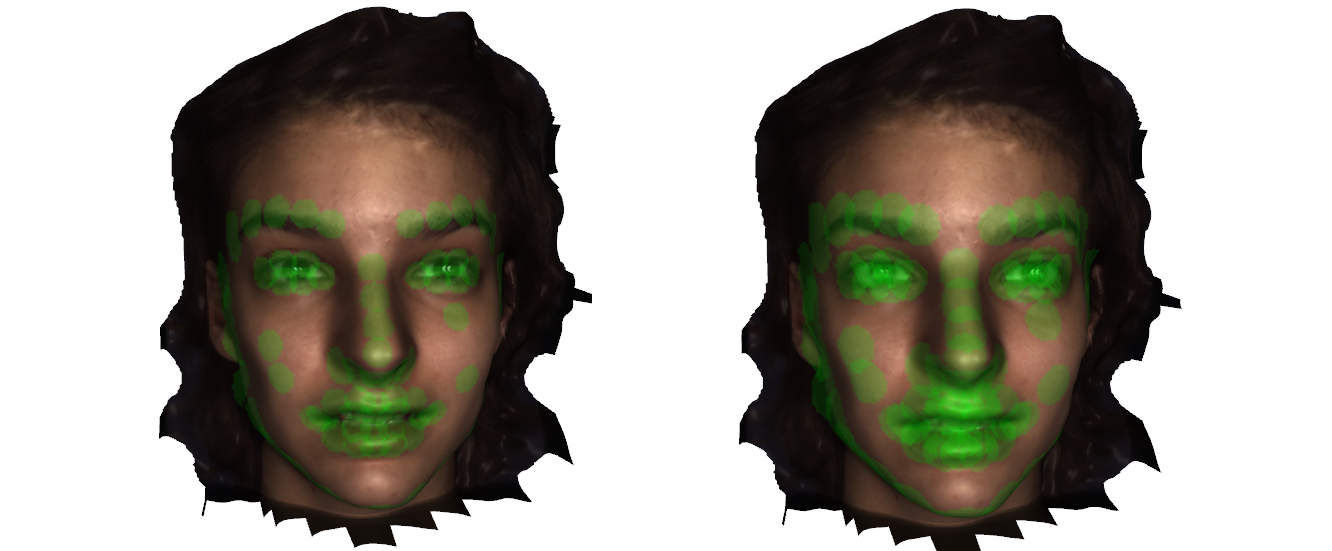}
    \caption{Extraction of local patches around the augmented landmarks. Left: Patches constructed with a neighborhood of 100 points around each landmark point. Right: Patches constructed with a neighborhood of 200 points.}
    \label{fig:patches}
\end{figure}

\subsection{Spatial-Temporal Graph Convolution Network (ST-GCN)}
\label{sec:st_gcn}
For each frame, let us denote $i$ each facial landmark corresponding to a vertex $v_i \in V$ in a graph $G=\{V,E\}$. Two types of edges are considered: spatial edges which connect landmarks based on a defined relationship between them and temporal edges which connect the same landmarks across consecutive frames. Once the spatio-temporal graph is generated, we employ a network which uses the generalization of regular convolutions for graphs, or else, \textit{graph convolutions}. Given a vertex $v_i$ and the sampling area $\mathcal{B}_i \subseteq V$ of $v_i$ which includes neighboring vertices $v_j$ of distance 1, a spatial graph convolution at an instant $t$ is defined as:

\begin{equation}
    f_{out}(v_i,t) = \sum_{v_j\in \mathcal{B}_i}{\frac{1}{Z_{ij}}f_{in}(v_j,t)\cdot w(l_i(v_j))},
    \label{eq:graph_conv}
\end{equation}

\noindent where $f_{in}$ and $f_{out}$ are the input and output feature maps, respectively, $Z_{ij}$ is a normalizing term which denotes the cardinality of the corresponding subset of $\mathcal{B}_i$, $w$ is the weight function and $l_i$ is a function which maps each vertex to the label of the corresponding weight vector. 

However, since graphs do not belong to the Euclidean space, adjacency matrices $\mathbf{A}$ and identity matrices $\mathbf{I}$ are needed to illustrate the landmark and self-connections, respectively. Therefore, the graph convolution can be now reformulated as: 

\begin{equation}
    \mathbf{f}_{out} = \mathbf{\Lambda}^{-\frac{1}{2}}(\mathbf{A}+\mathbf{I})\mathbf{\Lambda}^{-\frac{1}{2}} \mathbf{f}_{in} \mathbf{W},
\end{equation}

\noindent where $\mathbf \Lambda=[\Lambda^{ii}]_{i \in \{1,...,J\}}$ such that $\Lambda^{ii}=\sum_j(A^{ij}+I^{ij})$ and $\mathbf{W}$ is the weight matrix which consists of stacked weight vectors $w$. For spatio-temporal graphs of size $(C_{in},J,T)$, the resulting tensor has a dimension of $(C_{out},J,T)$, where $C_{in}$ and $C_{out}$ are, respectively, the number of input and output channels, $J$ is the number of spatial vertices (landmarks) and $T$ the temporal dimension. In practice, the graph convolution is applied by performing $m$ 2D convolutions on the graph vectors and multiplying the 2$^{nd}$ dimension of resulting tensors with the normalized adjacency matrix $\mathbf{\Lambda}^{-\frac{1}{2}}(\mathbf{A}+\mathbf{I})\mathbf{\Lambda}^{-\frac{1}{2}}$. 

The final step of the approach involves temporal convolutions on the output feature tensor $\mathbf{f}_{out}$. Temporal edges are essentially the trajectories of each landmark, therefore, a simple temporal convolution layer of kernel size $K$ is employed.

\section{Experiments}
\label{sec:experiments}

Our framework has been tested on the BU4DFE dataset, which consists of dynamic sequences of facial expressions. In this section, we present our results on 3D face recognition.

\subsection{Dataset}

For our experiments we used the \textbf{BU4DFE} dataset \cite{zhang2013high}. BU4DFE is a 3D dynamic facial expression database consisting of 101 subjects performing 6 facial expressions (anger, disgust, happiness, fear, sadness, and surprise). Each expression sequence contains about 100 frames resulting in an approximately 60,600 frame models. The 3D video sequences have a resolution of approximately 35,000 vertices each and the corresponding RGB videos have a frame resolution of about 1040x1329 pixels. 

\subsection{Evaluation Settings and Parameters}

Our implementation is based on the PyTorch ST-GCN \cite{yan2018spatial} code. Due to the different task and dataset specifications, we modified the network by keeping only 3 spatiotemporal graph convolutional blocks. We use 200 neighboring points per landmark as input features. These points consist of both RGB and shape features. Thus, the input channels per landmark sum up to 1200 after flattening the features. The Stochastic Gradient Descent optimizer is used with a decaying learning rate of 0.01.

Since BU4DFE dataset was not initially intended to be used for face recognition, we follow a cross-emotion protocol. We use three emotions for training and the rest for testing.  

\subsection{Face Identification Results}

In this section, we compare our approach to recent dynamic face recogntion methods, namely, Multi-Modal 2D-3D Hybrid Face Recognition (MMH 2D + 3D)~\cite{mian2007efficient}, Keypoint-based 3D Deformable Model (K3DM)~\cite{gilani2017dense}, Deep 3D Face Recognition Network (FR3DNet)~\cite{gilani2018learning}, 3D Deformation Signature~\cite{shabayek20203d} and 3D Sparse Deformation Signature~\cite{aouada20203d}. The obtained accuracy of recognition on BU4DFE dataset is reported in Table~\ref{tab:results}.

\begin{table}[ht!]
        \centering
        \caption{Accuracy of recognition on BU4DFE dataset. *Our approach was tested on cross-emotion setting.}
        \label{tab:results}
        {
        \begin{tabular}{| c | c | }
        
        \hline
            \textbf{Method} & \textbf{Accuracy} \\
             \hline
             MMH (2D + 3D)~\cite{mian2007efficient} & $94.20\%$\\
             \hline
             K3DM (3D)~\cite{gilani2017dense} & $96.00\%$\\
             \hline
             FR3DNet$_{FT}$ (3D)~\cite{gilani2018learning} & $98.00\%$\\
             \hline
             3DS (3D)~\cite{shabayek20203d} & $\mathbf{99.98\%}$\\
             \hline
             S3DS (3D)~\cite{aouada20203d} & $99.92\%$\\
             \hline
             \textbf{Ours}* & {${88.45\%}$}\\
             \hline
        \end{tabular}}
    \end{table}

Using the cross-emotion setting, our approach achieved a mean accuracy of 88.45\%, while the best performing approach reached 99.98\%. While the difference in accuracy reaches over 10\%, we argue that the splitting protocol used in our approach was more challenging.






\section{Conclusion}
\label{sec:conclusion}

In this paper, we proposed a novel framework for dynamic 3D face identification/recognition that relies on spatio-temporal graphs from facial landmarks. In order to enhance the discriminative power of the graphs, local textural and geometric features are extracted from the neighborhood around the corresponding landmarks. 

\section{Acknowledgements}
This work was funded by the National Research Fund (FNR), Luxembourg, under the project reference CPPP17/IS/11643091/IDform/Aouada.

{\small
\bibliographystyle{ieee_fullname}
\bibliography{egbib}
}

\end{document}